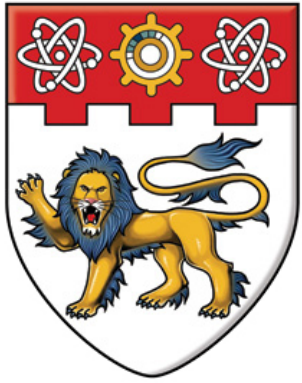
arXiv:1707.02412v1 [cs.LG] 8 Jul 2017CE/CZ4041: MACHINE LEARNING

**TRANSFER LEARNING RESEARCH PROJECT REPORT**
Application of Transfer Learning Approaches
in Multimodal Wearable Human Activity Recognition

Submitter By: CHEN HAILIN, CUI SHENGPING, SEBASTIAN LI

# Contents

# 1  Introduction

Machine learning has been the focus of global researchers for decades. A number of supervised machine learning techniques has been introduced, refined and used in a range of real world applications. Many existing methods can yield high performance within a certain data distribution and feature space. However, in real world applications, variations in data distribution and feature space can be introduced by many external factors, which leads to significant performance drops. To tackle this problem, an intuitive approach is to rebuild models tailored for certain data distribution and feature representation. Nevertheless, collecting a decent amount of high-quality training data for each case is unrealistic. Rebuilding models is neither an option as it is expensive in terms of time and computing power.

Transfer Learning is a research field which aims to improve learning on a certain task with knowledge transferred from related tasks. It allows different data distributions, tasks and representations between training and testing, and focuses on extracting knowledge that are shared between domains. The need for transfer learning arises from the above problems and various transfer learning techniques can be used in building a generalized model. More formally, transfer learning is defined as:

**Definition 1(Transfer Learning)**:
   Given a source domain $D_S$ and learning task $T_S$, a target domain $D_T$ and learning task $T_T$, transfer learning aims to help improve the learning of the target predictive function $f_T(.)$ in $D_T$ using the knowledge in $D_S$ and $T_S$ where $D_S \neq D_T$, or $T_S \neq T_T$

According to Pan (2010)[14], transfer learning and be subdivided into three fields: inductive transfer learning, transductive transfer learning and unsupervised transfer learning. The definitions are as follows:

**Definition 2 (Inductive Transfer Learning)**:
   Given a source domain $D_S$ and a learning task $T_S$, a target domain $D_T$ and a learning task $T_T$, inductive transfer learning aims to help improve the learning of the target predictive function $f_T(.)$ using knowledge in $D_T$ and $D_S$, where $T_S \neq T_T$.

**Definition 3 (Transductive Transfer Learning)**:
   Given a source domain $D_S$ with a learning task $T_S$, a target domain $D_T$ with a corresponding learning task $T_T$, transductive transfer learning aims to improve the learning of the target predictive function $f_T(.)$ in $D_T$ using the knowledge in $D_S$ and $T_S$, where $D_S \neq D_T$ and $T_S = T_T$. In addition, some unlabeled target-domain data must be available at training time.

**Definition 4 (Unsupervised Transfer Learning):**
   Given a source domain $D_S$ with a learning task $T_S$, a target domain $D_T$ and a corresponding learning task $T_T$, unsupervised transfer learning aims to help improve the learning of the target predictive function $f_T(.)$ in $D_T$ using the knowledge in $D_S$ and $T_S$, where $T_S \neq T_T$ and $Y_S$ and $Y_T$ are not observable.

With the above definition, in this research project, we focus on implementing methods of different approaches in transductive transfer learning. The task is the classification of human activity and the data set used is obtained from The Opportunity Challenge. We will describe and define the problem further in section 3.

## 2 Literature Review

With the definition from previous section, we now present on the various transfer learning techniques proposed in the research field. We will focus mainly on three approaches designed to solve transfer learning; 1)Instance based, 2)Feature based, 3)Parameter based. We will introduce the motivations, assumptions and mathematical definitions for each approach in the following section. Various methods in instance based and feature based approaches in transductive setting will be covered, and methods for parameter based approach will be mentioned in inductive transfer learning setting.

### 2.1 Instance Based Approaches

Instance based transfer learning is a class of transfer learning methods to resolve the differences in marginal probability distributions between source and target domain data. It aims to map the distribution of target domain from source domain by learning knowledge about inner structures of target and domain data. Basing on this knowledge, re-sampling and reweighing can be done to source data in order to improve performance in target domain.

One common approach in instance based transfer learning is inspired by importance sampling. The idea is to estimate the distribution of target domain using appropriate sampling of the source domain. Methods of this approach assumes that sufficient amount of unlabeled target data is available and that $P_S(y|x)$ is very similar to $P_T(y|x)$. If there are arbitrary $P_T(y|x)$ and $P_S(y|x)$, it is impossible to build a model that performs well in target domain only by distribution mapping. To simplify the estimation, the optimization problem can be written in the form of empirical risk minimisation:

$$\theta^* = \arg\min_{\theta \in \Theta} \sum_{(x,y) \in D_S} \frac{P(D_T)}{P(D_S)} P(D_S) l(x, y, \theta)$$

With the assumption $P_S(y|x) = P_T(y|x)$ and $P(x,y) = P(y|x)P(x)$, the equation can be written as:

$$\theta^* = \arg\min_{\theta \in \Theta} \sum_{i=1}^{n_S} \frac{P_T(x_{Ti})}{P_S(x_{Si})} l(x_{Si}, y_{Si}, \theta)$$

Therefore, the problem is to derive a proper estimation of the term $\frac{P_T(x)}{P_S(x)}$ denoted by $\beta(x)$, and it is often regarded as a *sample selection bias* problem by researchers. One solution is to train a classifier that outputs class membership probability estimates[14]. Huang et al. (2007)[2] proposed a reweighing method called Kernel Mean Matching (KMM). Instead of trying to recover the target distribution, KMM estimate $\beta_i := \beta(x_i)$ directly by kernel mean matching.

Let $\phi : X \to F$ be the mapping from $D$ to a a Reproducing Kernel Hilbert Space (RKMS) and $\mu$ denote the expectation operator: $\mu := E_{x \sim P_S(x)}(\phi(x))$. In Kernel Mean Mapping, $\beta$ can be derived by this minimization problem with constraints on $\beta$ distribution:

$$minimize_\beta \ || \ \mu(P_T) - E_{x \sim P_S(x)}[\beta(x)\phi(x)] \ ||$$

The problem is that $\mu$ is unknown and only a finite number of samples from $D_S$ and $D_T$ are available, and the empirical means would vary with the samples. Basing on the Central Limit Theorem, the precision limit can be obtained. Suppose samples of size $m$ and $m^{'}$ are drawn in an independent and identically distributed manner from $P_S(x)$ and $P_T(x)$ respectively. When $\beta(x) \in [0, B]$ is a fixed function, the sample mean $\frac{1}{m} \sum_i \beta(x_i)$ converges to a Gaussian with mean $\int \beta(x) dP_S(x)$ and deviation $\frac{B}{2\sqrt{m}}$.

The result is an empirical KMM optimization problem. The discrepancy between source and target means is minimized in the RKMS with constraints $\beta_i \in [0, B]$ and $|\frac{1}{m} \sum_{i=1}^{m} \beta_i - 1| < \epsilon$. And the discrepancy between means

$$|| \frac{1}{m} \sum_{i=1}^{m} \beta_i \phi(x_i) - \frac{1}{m'} \sum_{i=1}^{m'} \phi(x_i') ||^2$$

can be converted to a quadratic problem with kernels $K_{ij} := k(x_i, x_j)$ and $\kappa_i := \frac{m}{m'} \sum_{j=1}^{m'} k(x_j, x_j)$:

$$minimize_\beta \ \frac{1}{2} \beta^\mathsf{T} K \beta - \kappa^\mathsf{T} \beta$$

The paper also suggests a good choice of $\epsilon$ shall be $O(\frac{B}{\sqrt{m}})$ according to the deviation of $\beta(x)$

Dai et al. (2007) introduced a transfer learning version of AdaBoost[6] algorithm named TrAdaBoost[20]. The motivation is that when training a model for target domain, a subset of the source domain data is useful for training while others may degrade the performance due to the differences in distribution. TrAdaBoost assumes that the source and target domain data has the same set of features and labels, while the data distributions are different. Similar to AdaBoost, the algorithm assigns an initial weight $\omega_i$ to each data instance, repeatedly update these weights based on the error and retrain a new learner with the updated weights. However, in TrAdaBoost, the error is only calculated on target instances and the algorithm updates the weights of source instances in a different strategy from AdaBoost. TrAdaBoost uses source domain data and a small amount of labelled target domain data to train the model. Suppose in training data set $X$, $x_i \in D_S$ for $i = 1, ..., n$, while $x_i \in D_T$ for $i = n+1, ..., n+m$.

In each iteration, the error is calculated as:

$$\epsilon_t = \sum_{i=n+1}^{i=n+m} \frac{\omega_i^t \cdot | h_t(x_i) - c(x_i) |}{\sum_{i=n+1}^{n+m} \omega_i^t}$$

where $c(x)$ is the label for target instance x and $h(x)$ is the output of the learner given input x. Set

$$\beta_t = \frac{\epsilon_t}{1 - \epsilon_t} \quad and \quad \beta = \frac{1}{1 + \sqrt{2 \ln \frac{n}{N}}}$$

and then update the weight vector as:

$$\omega_i^{t+1} = \omega_i^t \beta^{|h_t(x_i) - c(x_i)|}, \ i = 1, ..., n$$

$$\omega_i^{t+1} = \omega_i^t \beta_t^{-|h_t(x_i) - c(x_i)|}, \ i = n+1, ..., n+m$$

As pointed in [20], TrAdaBoost can enhance the performance when the ratio between target training data and source training data is less than 0.1. However, since TrAdaBoost relies on single source data reweighing, it has poor performance when the correlation between source and target domain is weak[15]. In addition, another major issue of TrAdaBoost is that the source weights converge too fast for transfer to take place. Yao and Doretto (2010) proposed

Multi-Source TrAdaboost method (MSTrA), which aims to ensure that the training source data is relevant to the target domain. It combines multiple source domains as its training data and at each iteration, the most relevant domain is chosen to train the weak learner[21]. Samir and Chandan (2011) introduced dynamic cost factors into TrAdaBoost to address the convergence problem[16].

Jiang and Zhai (2007) proposed a framework of instance weighting in domain adaptation problems. Let the set of labelled source domain instances denoted by $D_s$, a large amount of unlabelled target domain instances $D_{t,l}$ and a small amount of labelled target domain instances $D_{t,u}$. The objective function is:

$$\hat{\theta} = \arg\max_{\theta}[\lambda_s \cdot \frac{1}{C_s} \sum_{i=1}^{N_s} \alpha_i \beta_i \log p(y_i^s | x_i^s; \theta) + \lambda_{t,l} \cdot \frac{1}{C_{t,l}} \sum_{j=1}^{N_{t,l}} \log p(y_j^{t,l} | x_j^{t,l}; \theta)$$

$$+ \lambda_{t,u} \cdot \frac{1}{C_{t,u}} \sum_{k=1}^{N_{t,u}} \sum_{y \in Y} \gamma_k(y) \log p(y | x_k^{t,u}; \theta) + log(p(\theta))]$$

where the $\gamma$ parameters control the parameter of each methods, $\beta$ is $\frac{P_T(x)}{P_S(x)}$, $\alpha$ indicates how close are $P_T(y|x)$ and $P_S(y|x)$, $\gamma$ shows the the confidence in the estimated probability $p(y|x_k^{t,u})$ and $log(p(\theta))$ to limit complexity. The framework basically summarized different methods in instance weighting with different kinds of data available.

## 2.2 Feature Based Approaches

Feature based transfer learning is trying to resolve the differences between source and target domain by changing the feature space. The big idea is that if the source and target data distribution has strong overlaps in certain feature space, then a discriminator trained from source domain in such feature space should also perform well for target domain in such feature space. According to Kouw[12], this approach can further be divided into two subcategories: Sample Transformation and Feature Augmentation. Blitzer[10] proposed a representative approach in Feature Augmentation, called Structural Correspondence Learning(SCL). As name suggested, it tries to identify correspondence between the source domain to the target domain by modelling how other features correlate to the pivot features. In SCL, pivot features are features that appear frequently and behave similarly in both domains. More formally, SCL algorithm is described in the figure above. As shown, SCL first removes m pivot features from the training data. It then

**Input:** labeled source data $\{(\mathbf{x}_t, y_t)_{t=1}^T\}$, unlabeled data from both domains $\{\mathbf{x}_j\}$

**Output:** predictor $f : X \to Y$

1. Choose $m$ pivot features. Create $m$ binary prediction problems, $p_\ell(\mathbf{x})$, $\ell = 1 \ldots m$

2. For $\ell = 1$ to $m$
   $$\hat{\mathbf{w}}_\ell = \underset{\mathbf{w}}{\operatorname{argmin}} \left( \sum_j L(\mathbf{w} \cdot \mathbf{x}_j, p_\ell(\mathbf{x}_j)) + \lambda ||\mathbf{w}||^2 \right)$$
   end

3. $W = [\hat{\mathbf{w}}_1 | \ldots | \hat{\mathbf{w}}_m]$, $[U\ D\ V^T] = \text{SVD}(W)$, $\theta = U_{[1:h,:]}^T$

4. Return $f$, a predictor trained on $\left\{ \left( \begin{bmatrix} \mathbf{x}_t \\ \theta \mathbf{x}_i \end{bmatrix}, y_t \right)_{t=1}^T \right\}$ constructs a binary classifier for each pivot feature; whose task is to predict whether this pivot feature occurs given a data instance. Thereafter, it trains m linear classifier to minimize the loss for predicting pivot feature occurrence. Next, Singular Vector Decomposition(SVD) is applied to the parameter matrix W of the above mentioned m classifiers to get matrix $\theta$, whose rows are essentially the top left singular vectors of W. Finally, we augment the input feature with $_i$ and input the augmented data to train any standard classifier.

Figure 1: SCL

SCL was first originally proposed as a general feature based technique and in particular; to resolve domain differences in PoS(part of speech) tagging problem. Ben-David[3] showed empirically that SCL can decrease the difference in both domains and decrease tagger error. However, it requires prior knowledge to choose pivot features. To decide which features select, Blitzer[3] uses the most frequently occurring words in PoS problem and proposed to use mutual information[8] in sentiment classification.

Instead of augmenting the original feature space, method in Sample Transformation aims to map the source and target domain into a share feature space where their difference is minimized. Most approaches in this category tries to minimize Maximum Mean Discrepancy(MMD) as it is non-parametric and it is also a suitable measure to compare probabilities in Reproducing Kernel Hilbert Space(RKHS). More specifically, MMD on transfer learning is defined as :

$$DIST(X'_S, X'_T) = || \frac{1}{n_1} \sum_{i=1}^{n_1} \phi \cdot \varphi(X_S) - \frac{1}{n_2} \sum_{i=1}^{n_2} \phi \cdot \varphi(X_T) ||_H$$

, where $\phi \in H$ and $\varphi$ is the projection map.
One representative method achieved this by dimensional reduction with the objective of minimizing MMD between source and target domains[19].It transforms the minimizing problem to semidefinite programming as:

$$\min_{\widetilde{K} \geq 0} trace(\widetilde{K}L) - \lambda trace(\widetilde{K})$$

s.t

$$\widetilde{K}_{ii} + \widetilde{K}_{jj} - 2\widetilde{K}_{ij} + 2\epsilon = d_{ij}^2, \forall (i,j) \in N, \widetilde{K} \cdot I = -\epsilon \cdot I$$

where

$$K = \begin{bmatrix} K_{SS} & K_{ST} \\ K_{TS} & K_{TT} \end{bmatrix} \in \mathbb{R}^{(n_1+n_2) \cdot (n_1+n_2)}$$

$$L_{ij} = \begin{cases} \frac{1}{n_1^2}, & \text{if } x_i, x_j \in X_S \\ \frac{1}{n_2^2}, & \text{if } x_i, x_j \in X_T \\ \frac{-1}{n_1 n_2}, & \text{otherwise} \end{cases}$$

$$K = \widetilde{K} + \epsilon I, \widetilde{K} \geq 0$$

After acquiring the kernel matrix $K$ from any standard semidefinite programming solver, this method applies Principle Component Analysis(PCA) on $K$ to transform original data representation $x_i$ to $x'_i$ for both domains. The results on synthetic data and real application showed a strong overlapping among both domains in the new reduced feature space.

Traditionally, feature based transfer learning usually involves feature learning, feature space transformation and discriminator learning. Recently, a deep learning based approach known as Domain Adversarial Neural Networks(DANN) incorporates the three processes together[22][13]. Its architecture is shown below. It consists of a feature extractor $G_f$, a domain classifier $G_d$ and a label predictor $G_f$, all of which are neural network based.

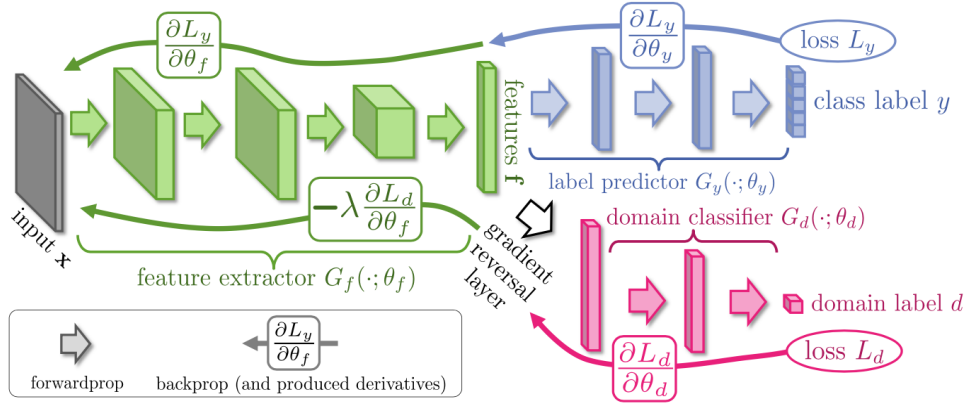

More formally, DANN aims to solve the following two equations to find:

$$(\hat{\theta}_f, \hat{\theta}_y) = \arg\min_{\theta_f, \theta_y} E(\theta_f, \theta_y, \hat{\theta}_d)$$

$$\hat{\theta}_d = \arg\min_{\theta_d} E(\hat{\theta}_f, \hat{\theta}_y, \theta_d)$$

where $\theta_f, \theta_y, \theta_d$ respectively represent the parameter for feature extractor, label predictor and domain classifier, and E is the functional defined as

$$E(\theta_f, \theta_y, \theta_d) = \sum_{i=1..N, d_i=0} L_y^i(\theta_f, \theta_y) - \lambda \sum_{i=1..N} L_d^i(\theta_f, \theta_d)$$

, where $L_y^i, L_d^i$ is respectively the loss of label predictor and loss of domain classifier for one input i, $d_i = 0$ denotes source domain data. By solving the above equations, we are able to find saddle points $\theta_f, \theta_y, \theta_d$, which minimize label predictor error and meanwhile maximize the domain classifier error. With the assumption of a good domain classifier, maximizing its loss by changing $\theta_f, \theta_y$, it essentially lets the feature extractor minimize the distribution difference between source and target domains. This will then improve label predictor under the assumption of covariate shift[17]. Fortunately, the two equations above can be solved by a series of Stochastic Gradient Descent updates as follows.

$$\theta_f \leftarrow \theta_f - \mu(\frac{\partial L_y^i}{\partial \theta_f} - \lambda \frac{\partial L_d^i}{\partial \theta_f})$$

$$\theta_y \leftarrow \theta_y - \mu \frac{\partial L_y^i}{\partial \theta_y}$$

$$\theta_d \leftarrow \theta_d - \mu \frac{\partial L_d^i}{\partial \theta_d}$$

## 2.3 Parameter Based Approaches

With the intuition that a model well trained from source domain data preserves some useful structures that can be applied to target domain, people develop parameter based approaches for transfer learning. This approach usually appears in the context of inductive transfer learning with the assumption that target domain task is related to source domain task. Although our research project belongs to transductive transfer learning, we include parameter-based approaches in the review since 1) it is an important approaches in transfer learning, 2) it is widely used in transfer learning with neural network in the form of layered transfer and 3)it can also be applied to boost our transfer learning performance.

The first generation of regularization-based methods from single-task to multitask learning is proposed by Evgeniou and Pontil (2004). The regularized multi-task learning approach utilizes task-coupling parameters to model the relation between tasks, thereby producing better performance[4]. Let $w_t$ denote the model parameter for task t. Hierarchical Bayesian methods assumes that all $w$ originates from a certain probability distribution, which means that all $w$ is related to a common $w_0$[1]. So for all $t$, $w_t = w_0 + v_t$, where $v_t$ is task specific and is less significant when the tasks are similar. To estimate $w_0$ and $v_t$ simultaneously, the optimization problem is defined as:

$$\arg\min_{w_0, v_t} \sum_{i=1}^{T} \sum_{i=1}^{N} l(x, y, w_t) + \frac{\lambda_1}{T} \sum_{t=1}^{T} \| v_t \|^2 + \lambda_2 \| w_0 \|^2$$

where $\lambda_1$ and $\lambda_2$ are regularization parameters. By changing the ratio between $\lambda_1$ and $\lambda_2$ we can impose constraints on how much each $w_t$ differs from one other.
In the context of transfer learning, the objective function can be rewritten as a framework for model-parameter based transfer learning[14]:

$$\arg\min_{\theta} \sum_{t \in S, T} \sum_{i=1}^{N} l(x, y, \theta_t) + \lambda_1 \| \theta \|^2 + \lambda_2 f(\theta)$$

In deep neural networks architecture, with tasks ranging from object recognition to speech recognition, people frequently discovered that the first few layers extract general features and the last few layers specialize those general features to task dependent prediction. Yosinski[9] conducted empirical experiments to test the transferability layers in neural networks. They defined the target domain as B and source domain as A, with XnB denoting that they train the model on domain X and freezing n number of layers starting with the first layer. Notation XnB+ is similar definition as XnB with variation that the first n layers are also able to be trained again(not frozen). The results are shown in the following plot.

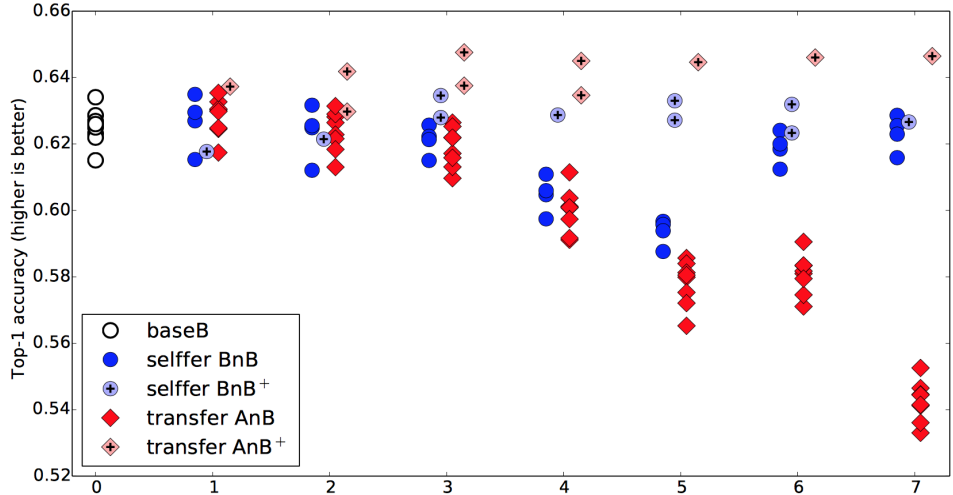

The most important result is AnB+ performance, which increases proportionally to the number of layers transferred. This reveals a surprising fact that all layers preserve useful parameters which are beneficial in training for target domain task.

Although this paper concludes such result in context of multitask learning, we believe layered transfer is also beneficial in transductive transfer learning, as our task in both domains is the same. This also meets the assumption that tasks in both domains are similar to each other.

## 3  Problem Description

In this section, we will describe the transfer learning problem t which we will apply different transfer learning approaches to solve. The dataset we use comes from the OPPORTUNITY challenge[7]. It is a dataset that contains recordings for Human Acitivity Recognition(HAR) in a sensor-rich environment[11]. In the dataset obtained from UCID, it contains the recordings of four different people, called subjects. Each subject's data includes 5 Active Daily Living(ADL) runs, in which a subject performs daily activities freely with a restriction that they are in a predefined activity set, and one drill run in which a subject follows specific instructions to perform 20 repetitions for each activity in the predefined activity set. This dataset contains both periodic and sporadic activity labels. In this report, we are mainly concerned with the task of recognizing sporadic activity as it is more difficult to predict and can thus, better show the improvement from transfer learning methods. The predefined sporadic activity set consists of 17 classes. The data is recorded with the frequency of 30 HZ and the total length of the dataset is 6 hours. This dataset originally has a lot of data instances being labelled as null, which according to paper[7] denote the state where the subject is not performing any meaningful activities. The following list shows statistics for different label classes:

| Class label | No. of Repetitions | No. of Instances |
|---|---|---|
| Open Door1 | 94 | 1583 |
| Open Door1 | 92 | 1685 |
| Close Door1 | 89 | 1497 |
| Close Door2 | 90 | 1588 |
| Open Fridge | 157 | 196 |
| Close Fridge | 159 | 1728 |
| Open Dishwasher | 102 | 1314 |
| Close Dishwasher | 99 | 1214 |
| Open Drawer 1 | 96 | 897 |
| Close Drawer 1 | 95 | 781 |
| Open Drawer 2 | 91 | 861 |
| Close Drawer 2 | 90 | 754 |
| Open Drawer 3 | 102 | 1082 |
| Close Drawer 3 | 103 | 1070 |
| Clean Table | 79 | 1717 |
| Drink from Cup | 213 | 6115 |
| Toggle Switch | 156 | 1257 |
| Null | 1605 | 69558 |

We define our transfer learning problem with this dataset here:

with $X_S, Y_S$ being labelled data of subject3(ADL1,2,3 and Drill) as source domain training data, with $X_{SV}, Y_{SV}$ being labelled data of subject3(ADL4,5) as source validation data, with $X_T$ being unlabelled data of subject4(ADL1,2,3 and Drill) as target domain training data, with $X_{TT}, Y_{TT}$ being labelled data of subject4(ADL4,5) as target domain testing data, we try to build a model which utilizes $X_S, Y_S, X_T$ to yield better performance in predicting $Y_{TT}$ from $X_{TT}$

# 4 Baseline DeepConvLSTM

In this section, we will introduce about our baseline method DeepConvLSTM including: 1.pre-processing of data 2.DeepConvLSTM architecture 3.Measure and Performance 4.Reasons for choosing DeepConvLSTM as baseline

## 4.1 preprocessing

As shown in the list above, null class accounts for 73% of all occurrences in all classes. This shows the dataset is highly imbalanced towards the null label. To better examine the predictor performance and transfer learning effect, we chose to exclude null class data in all training, validation and testing dataset.

In this study, we first conduct linear interpolation on the missing values in the sensor channels. We also normalize all channel data into the range of [-1,1]. As this dataset contains time series of data, we use a common method; sliding window with fixed length to segment the data. The window length is 24 data instances(0.8 seconds) and the sliding length is 12 data instances(0.4 seconds). After the data is segmented into consecutive windows, we assign the window with the label of the last data instance's as the ground truth label, which the following diagram shows[5].

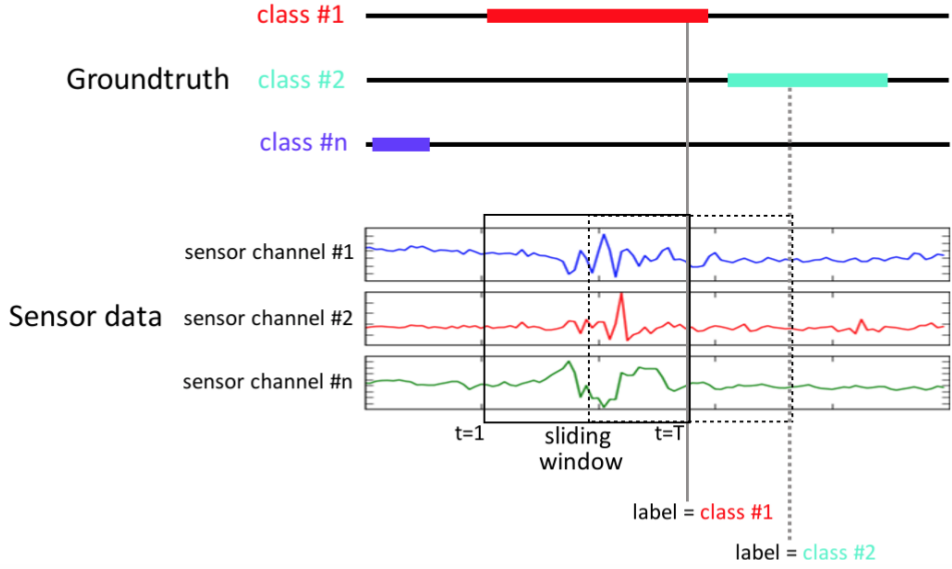

## 4.2 DeepConvLSTM architecture

DeepConvLSTM, as the name suggested, consists of convolutional layers and recurrent LSTM layers. As shown below, this model has 4 CNN layers in the beginning which convolve the input data on time axis. The first CNN layer convolve each sensor channel into 64 feature maps, while the following 3 CNN layers use $64 * 64$ kernel matrix to maintain 64 feature maps per channel. We set the convolution to be in valid mode to ensure the convolution kernel produce informed result in resulting feature map. In this way, a window input will be shrunk from 24 to 8 instances in the time dimension. This 8 time instances data is then fed into 2 LSTM layers, with each layer having 128 hidden nodes. These LSTM layers will output the prediction for each time instance. We then choose the output for the last instance as the input to the last softmax layer, to get the regularized probability prediction for each class.

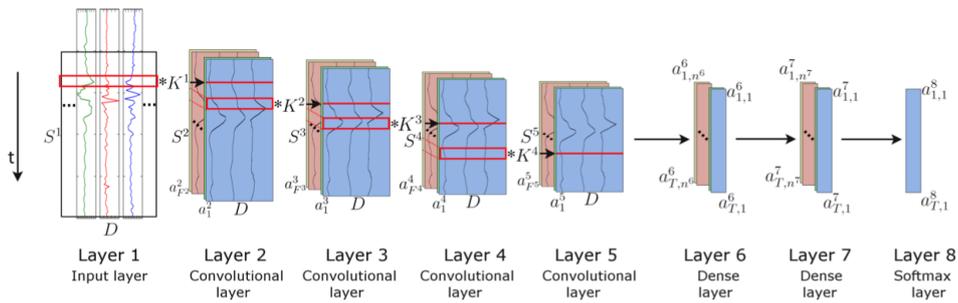

## 4.3 Measure and Performance

In this study, we deal with multi-label classification problem, with instances of each class label having different proportions. Thus, we define the measurement standard to be weighted F1 score, as proposed by Ordóñez[5]. Weighted F1 score is better than overall accuracy in dealing with an imbalance dataset. The weighted F1 score is defined as:

$$F_1 = \sum_i 2 * \omega_i \frac{precision_i \cdot recall_i}{precision_i + recall_i}$$

Using this measure, we conducted experiments on each of the subjects with ADL1,2,3, Drill as training data and ADL4,5 as testing data to test DeepConvLSTM performance. DeepConvLSTM results with other common methods from [7] are shown below.

| Method | S1 | S2 | S3 | S4 |
|---|---|---|---|---|
| LDA | 0.36 | 0.28 | 0.27 | 0.17 |
| QDA | 0.34 | 0.29 | 0.34 | 0.22 |
| NCC | 0.29 | 0.21 | 0.22 | 0.14 |
| 1NN | 0.56 | 0.53 | 0.58 | 0.46 |
| 3NN | 0.55 | 0.53 | 0.58 | 0.48 |
| DeepConvLSTM | 0.77 | 0.65 | 0.77 | 0.63 |

Based on the results, DeepConvLSTM outperforms all the commonly used baseline methods in all subjects.

### 4.4 Reasons in choosing DeepConvLSTM

In this study, we choose DeepConvLSTM mainly for two reasons 1)its performance is the best in the published methods in solving OPPORTUNITY dataset as a non-transfer learning problem 2)it has a comprehensive deep neural network structure. With regards to the first reason, we believe that a well performing baseline learner leaves greater potential for transfer learning to improve on target performance and it can reflect the effectiveness of the transfer learning techniques more objectively. As for the second reason, we observed that the DNN structure, especially with the initial CNN layers, is fairly suitable for parameter based transfer learning[9] and the recently published DANN[22] transfer learning method.

## 5 Methods Comparisons

In this section, we first apply our baseline learner to the transfer learning problem. Next we will apply different state of art methods to improve label prediction performance in target domain. We will then compare these methods' performance.

### 5.1 Baseline Performance

To get the predictions on $Y_{TT}$ and calculate f1 score, we apply models that are learned from $X_S$ to $X_{TT}$. We also list the result of the model learned from subject 4's data(ADL1,2,3, Drill) as the upper bound of our transfer learning solution.

| Training Data | Target F1 |
|---|---|
| Subject 3 | 0.38 |
| Subject 4 | 0.63 |

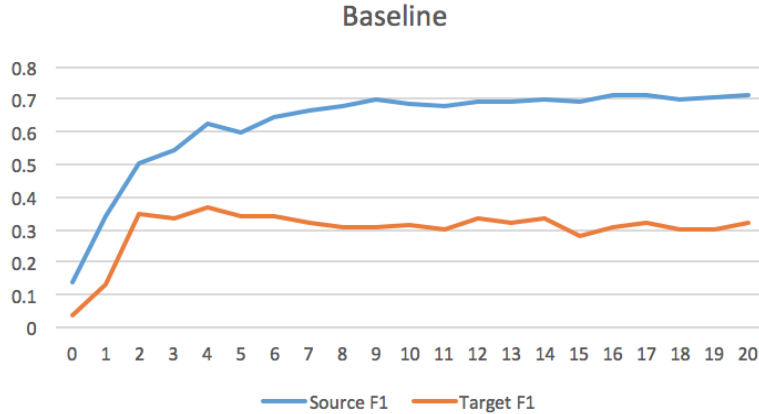

As shown in the list, the performance on target domain of a model learned from source domain is significantly lower than that of a model learned from target domain. The above graph shows the F1 score on source and target domain, with the increase of training iterations. The graph shows that the performance on target domain increases with source performance for the first 3 iterations, but then it drops and stabilizes at around 0.3 while the source domain accuracy keep increasing to over 0.7 in 21 iterations. Note that the performance is measured from validation set $X_{SV}, Y_{SV}$ and testing $X_{TT}, Y_{TT}$ set and that the model does not use any of those data for training. Thus, these two F1 score are objective performance indicators. The results above indicate that the transductive learning assumptions hold and there is a solid need for applying transfer learning methods in order to boost performance on target domain.

## 5.2 Instance Based(Loss Weighting)

Instance based transfer learning needs to estimate $\beta_i = \frac{P_T(x_i)}{P_S(x_i)}$ in order to optimize the model towards target distribution. In the various methods mentioned in the literature review, only a handful is applicable to deep neural networks. For example, in TrAdaBoost, training a series of neural network classifiers as learners is very expensive. For our instance-based approach, we map the distribution from source to target domain by assigning weights of loss to each instance. The weight of each instance is determined by a pre-trained probabilistic domain classifier. The domain classifier is trained using unlabelled source and target data $X_s, X_t$. Let $L_i$ denote the normalized output of probabilistic domain classifier of instance $x_i$, then we define:

$$\lambda_i = \frac{e^{\kappa \cdot L_i}}{C}$$

where $\kappa$ indicates the deviation between source and target distribution and $C$ controls the lower bound of loss, the loss function:

$$= \frac{1}{N} \sum_i \lambda_i \cdot D((S(\omega x_i + b), L_i)$$

This approach does not estimate $\beta$, instead it evaluates the loss based on how close the training instance is to target distribution. As shown in Figure 2, gradient descent is sluggish when the current training batch of data is vastly different from the target's, and the model is prevented from converging to source distribution too fast to transfer. When predicting in the target domain, this method achieves a F1 score of 0.4219, as shown is Figure 3.

Figure 2

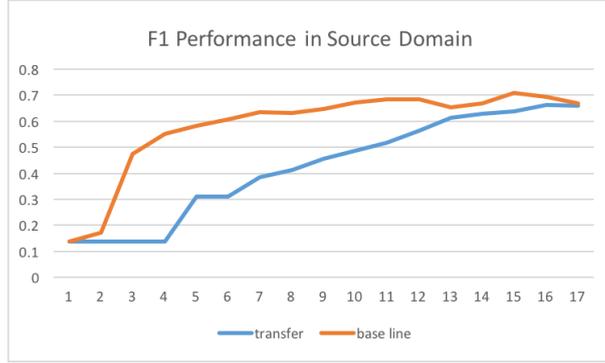

Figure 3

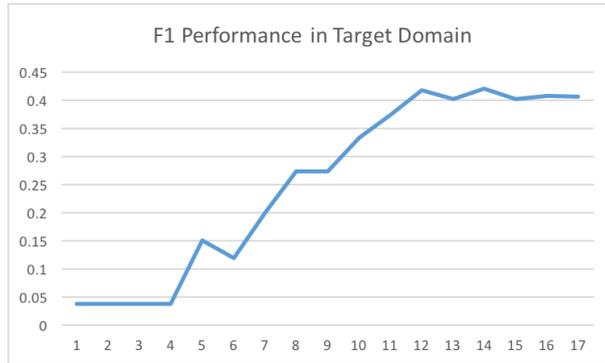

## 5.3 Feature Based(DANN)

Since our baseline learner is a DNN-based architecture, DANN[22] is the most suitable method in Feature based transfer learning for solving the HAR problem. More specifically, in DANN terminology, we treat the first two CNN layers as the feature extractor and the remaining layers as the label predictor. From this, we add another layer of with 128 LSTM nodes and a 2 output softmax layer as the domain classifier.

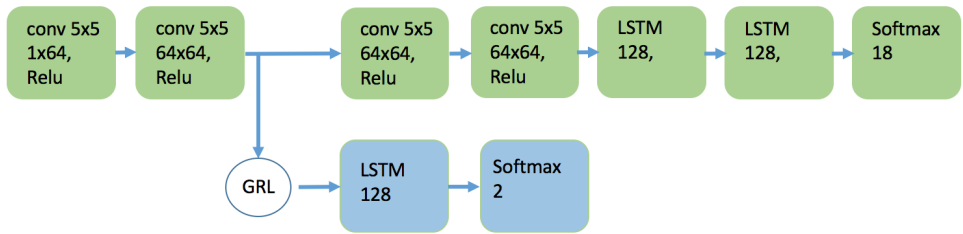

Note that 'GRL' denotes gradient reversal layer. According to gradient updates mentioned in section 2.2 DANN, $\lambda$ is multiplied with partial derivative as $\lambda \frac{\partial L_d^i}{\partial \theta_f})$. This is achieved in GRL layer in which the gradient is negated and multiplied by $\lambda$. As pointed out by Ganin[22], DANN is a generalized feature based method for DNN in Stochastic Gradient Descent(SGD). Although in his paper, Ganin chose to use Momentum optimizer, we use RMSProp optimizer in this model as 1.The optimizer performs the best according to Ordóñez[5] and 2.RMSProp is also an implementation of SGD. DANN integrates feature learning and training in the same neural network optimization process. However, choosing a suitable $\lambda$ to ensure the balance between domain invariance objective and label prediction objective is the key to make DANN beneficial. In Ganin's paper, he proposed the value of $\lambda$ to be

$$\lambda_p = \frac{2}{1 + \exp(-\gamma \cdot p)}, \gamma = 10, p = \frac{current\_iteration}{max\_iterations}$$

We conducted empirical runs to verify that this $\lambda$ setting is not applicable in our model. As the following graph shows, the accuracy for domain classifier increases to 90% in just two iterations. This indicates that the domain classifier effectively learns the differences while the feature extractor does not update itself to minimize the source and target differences efficiently. The orange line denotes the target domain's F1 score, which shows that this method performs almost the same as the baseline method and verifies the conclusion above.

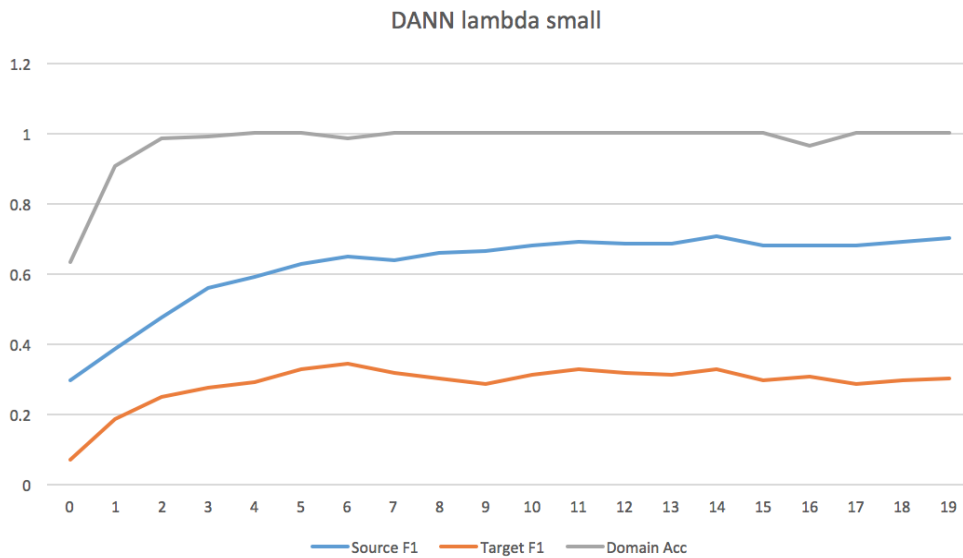

But this result also proves that our domain classifier is competent to learn and produce high quality prediction. We then propose an adaptive algorithm to enable the model by adjusting _lambda based on the performance of domain classifier. More specifically, the algorithm is defined as:

$$\lambda = \begin{cases} \alpha \cdot \lambda, \text{if} A_D > A_{Dmax} \text{ and } \lambda < \lambda_{max} \\ \frac{1}{\alpha} \cdot \lambda, \text{if } A_d < A_{Dmin} \text{ and } \lambda > \lambda_{mid} \\ \beta \cdot \lambda, \text{if } A_d < A_{Dmin} \text{ and } \lambda_{mid} > \lambda > \lambda_{min} \\ \lambda, \text{otherwise} \end{cases}$$

where $A_d$ is domain classifier accuracy in the current iteration. $A_Dmax$ and $A_Dmin$ are the domain accuracy upper and lower bound as defined by user; which we propose in our study to be 0.8 and 0.6. $\lambda_{max}, \lambda_{mid}$ and $\lambda_{min}$ are user defined parameters to prevent $\lambda$ from becoming too large or too small; which we defined in our study to be 10000, 10, 0.1. The $\alpha$ and $\beta$ are user defined parameters which control the growing/decreasing rate of $\lambda$, which in our case are defined as 1.5, 0.9 respectively. The result of using the above algorithm to update $\lambda$ is listed below.

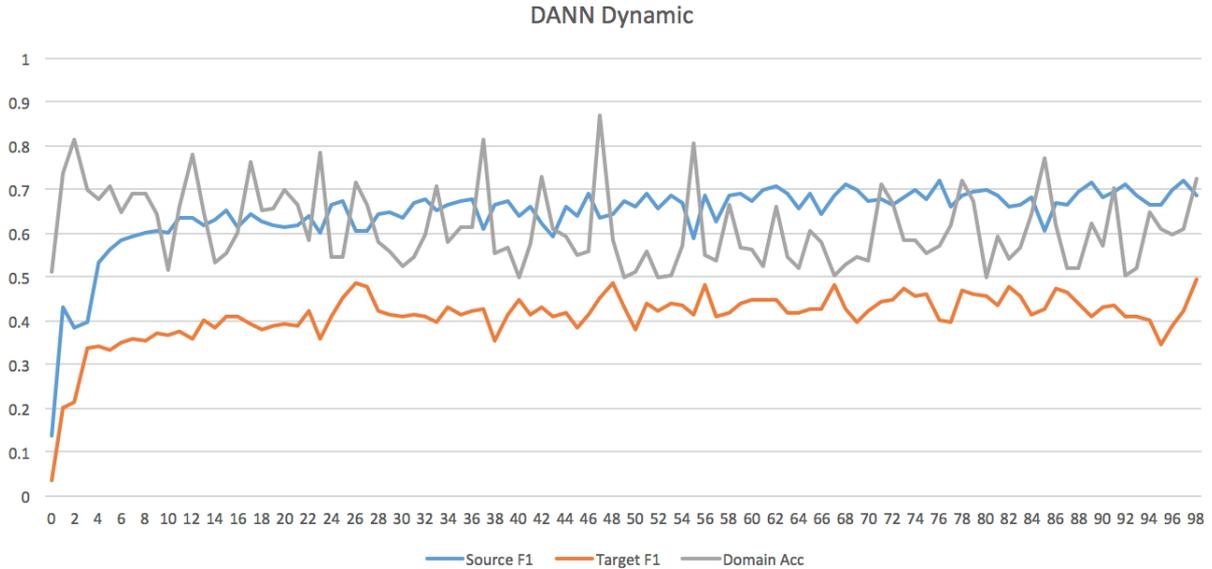

As shown by grey line above, with the updating of $\lambda$, domain classifier accuracy is regularized to be fluctuating from 50 percent to 80 percent. This is a good indication that the domain classifier is learning and the feature extractor is updating its parameters correctly to minimize the source and target domain's differences. The orange line denoting target domain F1 score shows that the model outperforms the baseline model, with the maximum F1 score achieved to be 0.49. This is a 29% improvement compared to the baseline model. It is worthwhile to note that DANN shows more variance in model performance compared to baseline model as the feature extractor is dynamically updated by both domain classifier and label predictor, with the weight $\lambda$ being dynamically updated. This explains why DANN maintains a comparatively high performance even after 100 iterations, while the baseline model performance drops after the first few iterations. As the feature extractor constantly updates itself and makes efforts to mix source and target domain, DANN is harder to overfit towards source domain, as compared to the baseline model.

## 5.4 Parameter Based(Layer transfer + Fine-tuning)

According to Yosinski[9], theoretically each layer of a deep neural network has captured some knowledge of the data that can be transferred to a new model domain. In real world applications, layered transfer, as a natural approach of transfer learning in neural network models, does not guarantee a performance gain. Neural network classifiers are discriminative in nature. Layered transfer needs to be able to identify layers that are specific to distribution and the model must be able to adjust itself to the target distribution.

Transfer learning with Convolutional Neural Network has proven to be effective in image recognition and natural language processing in recent researches[18]. The idea is that some CNN layers can be pre-trained to extract general features and then reused by tuning them towards a specific task/data distribution using a small amount of data. This can be very useful since one shortcoming of CNN is that it need large amount of data to train.

The problem definition we defined for parameter-based approach is different from the other two methods since labelled data is required for the model to learn the target distribution. The data available consists of the Drill set and three ADL sets of labelled source domain data $X_s$ as well as labelled target instances from one of the ADL sets $X_{t,l}$.

The method we used is layered transfer plus fine-tuning. The baseline model comprises of four convolution layers. We first train the whole model using $X_s$ and save the CNN parameters. In the transfer session, we restore the saved parameters and fine tune the model using the target data available. One concern is that when tuning the model with a small amount of target data, the model can easily overfit. Moreover, due to the small sample space, the target training data distribution can be biased. As discussed during literature review, general features are extracted in the first few CNN layers while the last few layers extract more specific features. With this consideration, we fixed the first few CNN layers, fine-tune the remaining layers and then use a low learning rate to prevent overfitting. During our experiments, we observed that when fixing all CNN layers, transferring actually produced performance worse than directly training on $X_{t,l}$. This proves that part of the layers captures features that are specific to distribution. The best performance is achieved by fixing the first two CNN layers and fine-tuning the last two layers (as in table 2) with a learning rate of 0.00005. The highest weighted F1 score is 0.5728 (tuned with ADL3) and the performance breakdown is shown in table 1. From the table we can observe that the performance in predicting each class is largely affected by the data distribution in the fine-tuning data.

Table 1: F1 Performance of Parameter-Based Approach

|  | $X_{t,l}$ | | |
|---|---|---|---|
| Class Label | ADL1 | ADL2 | ADL3 |
| 1 | 0.5200 | 0.5965 | 0.5161 |
| 2 | 0.6582 | 0.4262 | 0.0625 |
| 3 | 0.6370 | 0.5934 | 0.5972 |
| 4 | 0.6578 | 0.7317 | 0.5405 |
| 5 | 0.4444 | 0.5165 | 0.7516 |
| 6 | 0.5759 | 0.5369 | 0.4923 |
| 7 | 0.2857 | 0.1132 | 0.2041 |
| 8 | 0.1860 | 0.7647 | 0.5116 |
| 9 | 0.2857 | 0.2222 | 0.1961 |
| 10 | 0.2353 | 0.1290 | 0.2162 |
| 11 | 0.4167 | 0.2963 | 0.3500 |
| 12 | 0.4590 | 0.0667 | 0.3396 |
| 13 | 0.1714 | 0.3158 | 0.4255 |
| 14 | 0.2667 | 0.1778 | 0.5376 |
| 15 | 0.9351 | 0.9589 | 0.9733 |
| 16 | 0.8211 | 0.8356 | 0.8889 |
| 17 | 0.4828 | 0.5781 | 0.5333 |
| weighted | 0.5611 | 0.5561 | 0.5728 |

Table 2: A Comparison between Different Fixed Layers (tuned with ADL3 Data)

| Fixed CNN Layers | None | Layer 1 | 1,2 | 1,2,3 | 1,2,3,4 |
|---|---|---|---|---|---|
| Best F1 Score | 0.4822 | 0.5309 | 0.5728 | 0.5109 | 0.5071 |

## 5.5 Summary

In this section we have compared three representative methods in transfer learning, with our own modification to cater to our problem. The overall performance comparison is listed below.

| Method | Highest F1 |
|---|---|
| Baseline | 0.38 |
| Layer Transfer + Fine-tuning | 0.57 |
| Loss Weighting | 0.42 |
| DANN | 0.49 |
| Baseline with all target data | 0.63 |

With few labelled data in target domain, feature based approach produces a better result compared with the Baseline, DANN and Loss Weighting. The performance of instance based method depends on the correlation between source instances and target instances. There is a huge imbalance between 'useful' and 'useless' source instances, which is not properly handled by the current implementation. As for feature based method, DANN is designed for neural network models and it is less restricted by the correlation between domains, since it operates on higher level of feature spaces. Compared with the instance based approach, DANN shows better performance as expected. This is due to the fact that reducing source and target domain difference by mapping them both to a common feature space is generally easier and a far more flexible approach than by only weighting data instances. Our parameter-based approach has the best performance as expected since it needs labelled target data. One advantage of this approach is that with well-trained generalized CNN layers, tuning is rapid compared to other methods with pretty decent performance.

# 6 Roles of Members

| Name | Contribution |
|---|---|
| Chen Hailin | Implementing the baseline method and feature based method |
| Cui Shengping | Implementing instance based and parameter based method |
| Sabastian Li | Problem definition, presentation slides and data preprocssing |

# 7  Conclusion

Through this project, we researched on transfer learning methods and their applications on real world problems. By implementing and modifying various methods in transfer learning for our problem, we obtained an insight in the advantages and disadvantages of these methods, as well as experiences in developing neural network models for knowledge transfer. Due to time constraint, we only applied a representative method for each major approach in transfer learning. As pointed out in the literature review, each method has its own assumptions, strengths and shortcomings. Thus we believe that an ensemble-learning approach combining the different methods should yield a better performance, which can be our future research focus. We have uploaded all source code used in this research project and we are open to discussion on topics involved in this report.